\documentclass{article}
\usepackage[preprint]{neurips_2026}
\usepackage[utf8]{inputenc}
\usepackage[T1]{fontenc}
\usepackage{hyperref}
\usepackage{url}
\usepackage{booktabs}
\usepackage{amsmath}
\usepackage{graphicx}
\usepackage{xcolor}

\title{Whose Gold? Annotator-Pool Disagreement Is Large at the Item Level,\\and Hidden by Small Leaderboards}

\author{%
  Anik Jha\\
  Independent Researcher\\
  \texttt{anik.k.jha@gmail.com}%
}

\begin{document}
\maketitle

\begin{abstract}
Preference benchmarks are built by hiring annotators, and the identity of those
annotators is treated as an implementation detail. We measure what that detail
buys. On the $2{,}885$ \textsc{MultiPref} items where both pools are internally
unanimous, so no tie-breaking convention is consulted at all, expert and crowd
annotators assign a different majority label to $23.6\%$ and name the
\emph{opposite winner} on $9.2\%$; on the $246$ comparably unanimous \textsc{MT-Bench} cells, benchmark authors
and recruited experts differ on $30.5\%$ and reverse on $8.5\%$. Yet on both corpora the
resulting model leaderboards are \emph{bit-identical}: Kendall $\tau = 1.00$ with
zero of six models displaced.

That invariance is far weaker evidence than it looks, and we quantify how weak. Switching
pools moves a model's win rate by $1.9$pp ($\mathrm{SD}$), one adjacent pair in our own
leaderboard sits $0.8$pp apart and had a $38\%$ chance of swapping, and an item-level
bootstrap displaces at least one model in $28\%$ of resamples. The observed zero is the
common outcome, not a property of aggregation: on the same measured perturbation, a
ten-model leaderboard is displaced with probability $0.86$ and a twenty-model leaderboard
with probability $0.9997$. Reporting a six-model leaderboard is safe; the safety does not
generalise, and everything that consumes labels \emph{per item} is not safe at any size.
We make the distinction precise, show that a widely used dataset's stated assumption of no
intra-group annotator variability is false, and show that an LLM judge tracks the crowd
pool over the expert pool on all three models we test, including one from a different
vendor. All code, per-call outputs, and pre-registered decision rules will be released upon
acceptance.
\end{abstract}

\section{Introduction}

Every preference benchmark rests on a hiring decision. Someone chose whether to
recruit crowdworkers or domain experts, how many to put on each item, and how to
collapse them into a label. That choice is normally reported in an appendix and
then never mentioned again, because the field's implicit model is that annotators
are noisy instruments measuring a single underlying quantity.

The dataset we study states this assumption explicitly. \textsc{MultiPref}
\citep{miranda2024hybrid} deliberately collects two annotator pools (ordinary
crowdworkers and screened domain experts) and its Limitations section says:
\emph{``One of our key assumptions is that there is no variability in intra-group
annotators \ldots{} the dataset disambiguates between normal and expert
crowdworker annotations. We leave this exploration for future work.''} This paper
is that exploration, and the assumption does not survive it.

A human-agent team is scored against exactly these labels. When a team's output is
graded, or a monitor is validated against ``human agreement'', the gold standard is
whichever pool was hired; if that choice moves half the item-level decisions, it
moves the measured competence of the team as much as the team does. That is the
sense in which this is a question about human-AI coevolution rather than about
dataset hygiene.

Our contribution is a measurement and a distinction:

\begin{enumerate}\itemsep2pt
\item \textbf{Item-level divergence is large.} Two pools annotating the same
items reverse the winner on $9.2\%$ (\textsc{MultiPref}, on the $2{,}885$ items
needing no tie-breaking convention) and $8.5\%$ (\textsc{MT-Bench}, likewise), and
change the majority label on $23.6\%$ and $30.5\%$. On \textsc{MultiPref} the two \emph{experts} on a single item disagree
with each other $50.1\%$ of the time.
\item \textbf{Little of it reaches \emph{this} leaderboard, and that is a statement about
spacing.} Ranking the same models under expert-gold and crowd-gold labels gives
$\tau = 1.00$ on both corpora, with no model displaced. But the pool perturbs win rates by
enough that displacement should be expected once a leaderboard holds more than a handful
of models (\S\ref{sec:threshold}), so the aggregate result bounds nothing about the arenas
the field actually publishes.
\item \textbf{The distinction that follows.} Reporting a leaderboard is safe.
Reward-model training, active-annotation routing, and LLM-judge validation all
consume \emph{per-item} labels. We measure one of these and \emph{conjecture} the
others: the mechanism is shared, but only the judge is tested here.
\item \textbf{One such consumer, measured.} Three LLM judges, two from one vendor and
one from another, all agree measurably more with the crowd majority than the expert
majority ($-6.9$, $-5.2$ and $-3.7$pp, all three CIs excluding zero). ``Our judge agrees with humans
$X\%$ of the time'' is therefore partly a statement about the hiring decision.
\end{enumerate}

\section{Setup}

\textbf{Corpora.} \textsc{MultiPref} \citep{miranda2024hybrid} contains $10{,}461$
pairwise comparisons over six models, each annotated twice by crowdworkers and
twice by screened experts on a five-point preference scale. \textsc{MT-Bench}
human judgments \citep{zheng2023judging} contains $3{,}355$ human votes; $292$
item cells are rated by both the benchmark's own authors (\texttt{author\_*}) and
recruited expert annotators (\texttt{expert\_*}), giving an independent pool
contrast over a different task and a different six-model set.

We attempted two further corpora and report the failures, because they explain why
this went unmeasured. \textsc{PRISM} \citep{kirk2024prism} assigns one participant
per conversation, so no two annotators ever rate the same item and the contrast is
undefined. \textsc{HelpSteer2} \citep{wang2024helpsteer2} ships aggregated
per-response attribute scores with no annotator identity and no pairwise
preferences. Multi-annotated preference corpora that retain per-annotator labels
are scarce.

\textbf{Aggregation.} For each pool we take the majority label per item and score
a model by its mean credit over the comparisons it appears in. When a pool is split
we force the tie category, but the primary analysis avoids the question entirely
(see below). Rank comparisons use Kendall $\tau$ with a
\emph{paired} bootstrap: each resample draws the same item indices for both label
sources, since the two rankings differ only in whose labels are used.

\textbf{Pre-registration.} Decision rules were written into a versioned plan
before each run. The leaderboard analysis carried the rule \emph{``kill if
$\tau > 0.9$ and the CI lower bound $> 0.9$; report the null and stop, do not
enlarge the model set hunting for a flip.''} We report the outcome under that rule
rather than a post-hoc one.

\paragraph{Tie-breaking is not a detail here, so we avoid it.}
Each pool has two annotators and the two experts on an item disagree $50.1\%$ of the
time, so a ``pool majority'' is decided by convention on roughly half the corpus, and
the corpus is skewed $1.78{:}1$ toward B. Our primary analysis therefore consults no
convention: we restrict to the $2{,}885$ items ($27.6\%$) where \emph{both} pools are
internally unanimous. That conditions on the easier items, so $9.2\%$ is a floor, not
an unbiased estimate; the bracket across all conventions is $2.5\%$--$29.8\%$.
Table~\ref{tab:tiebreak} gives the sensitivity. Our leaderboard ran a rule ranking A
over B over Tie. Four of five conventions, including the one we ran, leave the
ranking at $\tau = 1.0000$: under our own rule, $29.8\%$ of
items reverse and the leaderboard still does not move. One hypothetical rule (A over
Tie over B) does break it ($\tau = 0.467$); its mirror does not, which identifies it
as an artifact of the B-skew rather than a property of the annotators.

\section{Divergence at the item level, invariance in aggregate}

\begin{figure}[t]
\centering
\includegraphics[width=0.96\textwidth]{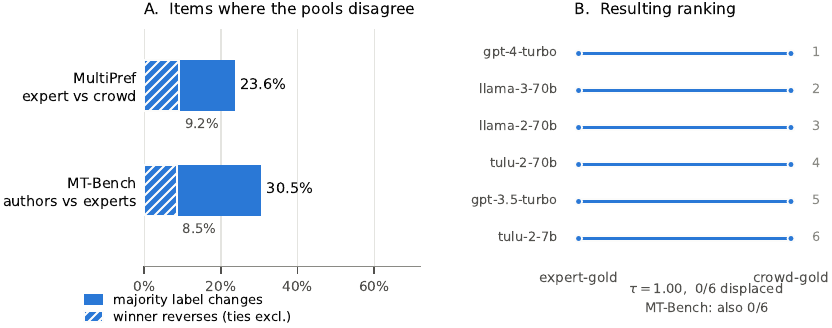}
\caption{Two pools annotating the same items disagree substantially
(\textbf{A}) and produce identical model rankings (\textbf{B}). Hatching marks
outright winner reversals, a strict subset of majority-label changes. Panel B
shows \textsc{MultiPref}; \textsc{MT-Bench} likewise displaces zero of six. Panel B is a
six-model leaderboard, and \S\ref{sec:threshold} shows that is doing much of the work:
the same perturbation displaces a ten-model board with probability $0.86$.}
\label{fig:main}
\end{figure}

\begin{table}[t]
\centering
\small
\caption{The same pattern on two corpora with different pool contrasts. Winner
reversal excludes ties on both sides, so it counts only items where both pools
name a decisive and opposite winner. \textsc{MultiPref} is restricted to items where both pools are
internally unanimous, and \textsc{MT-Bench} likewise, so no tie-breaking convention
enters either corpus.}
\label{tab:main}
\begin{tabular}{lccccc}
\toprule
Corpus & Pool contrast & Items & Majority differs & Winner reverses & Kendall $\tau$ \\
\midrule
\textsc{MultiPref} & expert vs.\ crowd    & $2{,}885$  & $23.6\%$ & $9.2\%$  & $1.0000$ \\
\textsc{MT-Bench}  & authors vs.\ experts & $246$      & $30.5\%$ & $8.5\%$  & $1.0000$ \\
\bottomrule
\end{tabular}
\end{table}

\begin{table}[t]
\centering
\small
\caption{Every tie-breaking convention we tried, on all $10{,}461$ \textsc{MultiPref}
comparisons. The first row is the rule our leaderboard actually used. Only the last,
a hypothetical rule ranking A over Tie over B, moves the ranking; its mirror image
does not, identifying it as an artifact of the corpus's $1.78{:}1$ B-skew.}
\label{tab:tiebreak}
\begin{tabular}{lccc}
\toprule
Tie-break rule & Majority differs & Winner reverses & Kendall $\tau$ (displaced) \\
\midrule
A $>$ B $>$ Tie (what we ran)   & $44.9\%$ & $29.8\%$ & $1.0000$ ($0/6$) \\
Both pools unanimous (no rule)  & $23.6\%$ & $9.2\%$  & $1.0000$ ($0/6$) \\
Force ties to Tie               & $41.2\%$ & $2.5\%$  & $1.0000$ ($0/6$) \\
B $>$ Tie $>$ A (mirror)        & $36.6\%$ & $9.4\%$  & $1.0000$ ($0/6$) \\
A $>$ Tie $>$ B (hypothetical)  & $49.8\%$ & $16.8\%$ & $0.4667$ ($4/6$) \\
\bottomrule
\end{tabular}
\end{table}

Table~\ref{tab:main} and Figure~\ref{fig:main} give the result. On
the convention-free \textsc{MultiPref} subset, one comparison in eleven has expert and crowd majorities naming
opposite winners. The divergence is not concentrated on hard cases: bucketing
items by the quality gap between the two models being compared gives $48.1\%$,
$50.3\%$ and $51.4\%$ majority divergence across tertiles (Pearson $r = +0.030$) for
closely matched through widely separated
ones. It is flat.

The aggregate is untouched. Both leaderboards are identical, $\tau = 1.0000$, zero
of six models displaced, on both corpora. The \textsc{MultiPref} bootstrap CI is
$[0.867, 1.000]$ and the \textsc{MT-Bench} CI is $[0.733, 1.000]$; the intervals
are wide because six models admit few discordant pairs, so we claim only that no
displacement occurs, not that none could.

A $9.2\%$ item-level reversal rate that cancels in a mean is still a $9.2\%$ reversal rate
for any consumer that does not take the mean. But before drawing comfort from the
cancellation, it is worth asking how much comfort six models can supply.

\section{How much invariance is that, exactly?}\label{sec:threshold}

A leaderboard can be invariant because aggregation is robust, or because nothing on it was
close enough to swap. Those are different claims and a $\tau$ of $1.00$ does not separate
them, so we measure the second directly.

\textbf{The perturbation.} Switching pools moves each model's win rate: across our six
models the crowd-minus-expert shift ranges from $-2.1$pp to $+2.8$pp, with
$\mathrm{SD} = 1.9$pp. A swap between two adjacent models needs the \emph{difference} of
their two shifts to exceed the gap between them, and that difference has
$\mathrm{SD} = 2.6$pp. Any two models closer together than about $4.3$pp therefore carry at
least a $5\%$ chance of trading places when the annotator pool changes.

\textbf{Our own leaderboard was lucky.} Its adjacent gaps are $11.1$, $2.9$, $2.8$, $0.8$
and $2.5$pp. The $0.8$pp pair had a $38\%$ probability of swapping and did not. Resampling
items and recomputing both leaderboards puts the whole-leaderboard displacement rate at
$28\%$: the observed zero is the modal outcome, not a reliable one. A parametric estimate
from the measured perturbation alone gives $28.1\%$ for a six-model board, against the
$28.2\%$ the bootstrap returns, so the two routes agree closely enough to extrapolate from.

\textbf{Where the invariance ends.} Holding the perturbation fixed and spreading $K$ models
uniformly over the same win-rate span, the probability that at least one is displaced is
$0.28$ at $K=6$, $0.86$ at $K=10$, $0.9997$ at $K=20$, and indistinguishable from one at
$K=50$ (Table~\ref{tab:threshold}). Real arenas cluster models more tightly than uniform
spacing, so these are lower bounds.

\begin{table}[t]
\centering
\small
\caption{Probability that changing the annotator pool displaces at least one model, as a
function of leaderboard size, holding the measured perturbation ($\mathrm{SD} = 2.6$pp on
the pairwise difference) fixed and spreading $K$ models uniformly across the observed
win-rate span. The $K=6$ row is reproduced independently by an item-level bootstrap of the
real corpus ($0.282$), which is why we are willing to read the rest of the column.}
\label{tab:threshold}
\begin{tabular}{lccccc}
\toprule
Models on the leaderboard & 6 & 10 & 20 & 30 & 50 \\
\midrule
Gap between adjacent models & $4.0$pp & $2.2$pp & $1.1$pp & $0.7$pp & $0.4$pp \\
P(at least one displaced)   & $0.28$  & $0.86$  & $0.9997$ & $1.00$ & $1.00$ \\
\bottomrule
\end{tabular}
\end{table}

This changes what the aggregate null licenses. It is not that leaderboards are robust to
who annotates. It is that a six-model leaderboard with one large gap in it survived, and
that the same measured perturbation would move a leaderboard of the size the field
actually publishes. The reassuring reading of $\tau = 1.00$ is available only at small $K$.

\section{What inherits the divergence}

The most widely deployed per-item consumer is the LLM judge, validated by
statements of the form \emph{``our judge agrees with humans $X\%$ of the time.''}
If a judge tracks one pool more than the other, $X$ is partly a statement about
the hiring decision.

We judged a stratified sample of \textsc{MultiPref} with three local open-weight
models, in both presentation orders, under a schema-constrained five-point output
identical to the human scale. All three agree measurably more with the crowd majority than
with the expert majority (Table~\ref{tab:judge}). The third is from a different vendor and
a different pretraining lineage, and is served at the same Q8\_0 precision as the other
two, so vendor is the only thing that changes; it shows the same sign at a smaller
magnitude. On pool-split
items (those where the two majorities disagree), Qwen3.6-35B-A3B matches the
crowd on $46.3\%$, the expert pool on $34.0\%$ and neither on $19.7\%$;
Qwen3.6-27B gives $40.6\%$, $32.7\%$ and $26.6\%$; Gemma-4 gives $39.5\%$, $33.3\%$ and
$27.2\%$. Among the items where the judge matched \emph{either} pool the crowd share is
$57.6\%$, $55.4\%$ and $54.3\%$.

We flag one honest complication. An initial run of the second model at half this
sample size did \emph{not} clear the pre-registered replication bar
($-3.2$pp, CI $[-7.3, +1.2]$); the effect appeared only once power was matched,
and we report that rather than presenting the matched run as the first attempt. We
had also explained the effect as attenuation by label noise, which predicts a
cleaner judge shows a \emph{larger} effect. The data run the other way, and now do so
across all three judges: order-swap direction reversal falls $44.6\% \to 24.5\% \to
23.7\%$ while the crowd lean falls $6.9 \to 5.2 \to 3.7$pp. The cleaner the judge, the
\emph{smaller} the effect. That explanation is therefore wrong and we withdraw it; the
mechanism remains open, and this monotone pattern across three models is the sharpest clue
we can offer toward it. One candidate we have not tested: post-training preference data
for open-weight models is itself typically collected from paid crowd annotators rather
than screened experts, so a judge's own alignment could imprint a crowd-shaped prior
independent of the comparison task. We do not have alignment-data provenance for any of
the three judges and cannot test this without it, so it remains a hypothesis, not a finding.

\begin{table}[t]
\centering
\small
\caption{LLM-judge alignment with each pool on \textsc{MultiPref}, on items where
the judge is self-consistent across presentation orders, within each model's pre-registered 2{,}500-item stratified sample. Negative difference means
the judge agrees more with the crowd majority. All three clear the
pre-registered bar, which is conjunctive over agreement and erasure. Under the
committed primary estimator (midpoint order-combination, all items rather than the
self-consistent subset) the agreement difference is $-0.026$, $-0.043$ and $-0.022$ for
the three judges in table order; the two whose agreement CI touches zero survive the
conjunctive bar on erasure ($+2.3$pp and $+2.8$pp), together with the self-consistent
subset shown here. Every judge is negative under both estimators.}
\label{tab:judge}
\begin{tabular}{lcccc}
\toprule
Judge & Items & Agree (expert) & Agree (crowd) & Difference [95\% CI] \\
\midrule
Qwen3.6-35B-A3B (Qwen)   & $1{,}383$ & $0.469$ & $0.538$ & $-0.069$ $[-0.103, -0.034]$ \\
Qwen3.6-27B (Qwen)       & $1{,}884$ & $0.427$ & $0.479$ & $-0.052$ $[-0.082, -0.022]$ \\
Gemma-4-26B-A4B (Google) & $1{,}904$ & $0.432$ & $0.469$ & $-0.037$ $[-0.067, -0.008]$ \\
\bottomrule
\end{tabular}
\end{table}

\paragraph{Two further measurement choices move the judge more than the pool does.}
Both are by-products of the judge arm rather than targets of it, and both are larger than
the $5$--$7$pp pool effect the section is about, which is the reason to record them. First, all three
judges are strongly position sensitive, reversing direction under order swap on
$44.6\%$, $24.5\%$ and $23.7\%$ of items, consistent with published position-bias
measurements \citep{positionbias2026} and, importantly, differing by a factor of
$1.8$ between the two Qwen models despite being the same family. Second, holding prompt, temperature
and seed fixed and changing only the output format (a JSON object versus a bare
constrained token), the same judge agrees with itself on only $47.2\%$ of items
$(n=500)$, with a systematic shift in one direction ($146$ items move A$\to$B, $3$
move B$\to$A). Judge numbers are conditional on output format, which is rarely
reported \citep{judgesense2026}. We can localise the cause: re-scoring the same
items with the same model at bf16 (rather than the Q8\_0 quantisation the sweep
served) by \emph{prefilling} the JSON scaffolding and reading the next-token
distribution recovers $94.6\%$ agreement with the generated labels, against
$47.2\%$ for a bare digit. The $5.4\%$ residual absorbs the quantisation change as
well as the decoding path, so it is an upper bound on both and the attribution to
scaffolding is conservative. The scaffolding tokens move the judgement, not
the decoding path.

\section{What to do instead}

\textbf{Report the pool.} An LLM-judge agreement figure without the annotator pool
that defines its reference labels is under-specified by roughly the size of the
effect being reported.

\textbf{Do not infer per-item reliability from leaderboard stability.} These are
different quantities and our data separates them by a wide margin. Nor should
leaderboard stability be inferred from a leaderboard: at six models it is largely a
statement about how far apart the models were, and \S\ref{sec:threshold} gives the
arithmetic for checking whether a given board is in the safe regime.

\textbf{Counterbalance order, always.} At $24.5$--$44.6\%$ direction reversal, a
single-order judge measurement on this task is close to a coin flip.

\textbf{Keep per-annotator labels.} Two of the four corpora we examined cannot
support this analysis at all, purely because they discarded annotator identity.

\section{Limitations}

Both corpora rank only six models, and \S\ref{sec:threshold} measures rather than
asserts what that costs: the aggregate invariance is a statement about this leaderboard's
spacing and does not transfer to larger arenas. The extrapolation in
Table~\ref{tab:threshold} assumes the pool perturbation is independent across models and
approximately Gaussian; it reproduces the $K=6$ bootstrap closely, but we have six models
with which to check it, and a corpus with more models is the obvious way to test it
properly. The \textsc{MT-Bench} contrast rests on $246$ convention-free cells of $292$
doubly-rated ones. The judge arm now spans two vendors: \textsc{Laguna S 2.1} does not load in our stack, and
a substitute prefill path reproduced the generation path on $94.6\%$ of items against a
pre-registered $95\%$ bar, so we ran a third judge from a different pretraining lineage
instead, at the same Q8\_0 precision as the other two. The crowd lean replicates across all
three, and the two smallest effects belong to the two cleanest judges. Three models from two
vendors is still not a claim about judges in general, and the effect we can bound is between
roughly $4$ and $7$pp. We report the null, the failed under-powered run, and the withdrawn
explanation alongside the positive results deliberately: the paper's claim is
about measurement validity, and it would be self-undermining to present it
selectively. All local judge inference ran on a single on-premises NVIDIA GB10
(Grace Blackwell) workstation serving Q8\_0-quantised GGUF checkpoints; the three
judge sweeps (2{,}500 comparisons $\times$ 2 presentation orders each) took
49--122 minutes wall-clock per model.

\section{Related work}

Perspectivist and disagreement-aware evaluation argues that annotator
disagreement is signal rather than noise \citep{perspectivist2026} and proposes
aggregation that models annotator confusion \citep{stableval2026}. Those methods
rank \emph{systems} using human annotators; we ask the prior question of whether
the choice of annotator population changes the answer, and we measure LLM judges
against each population separately. Work on LLM-judge bias documents position,
verbosity and format effects \citep{positionbias2026, judgesense2026}; our
position and format numbers replicate that line rather than extend it, and are
reported here as controls on the pool measurement.

\bibliographystyle{plainnat}

\end{document}